\documentclass{article}
\usepackage{arxiv}

\usepackage{amsmath} 

\usepackage[utf8]{inputenc} 
\usepackage[T1]{fontenc}    
\usepackage{hyperref}       
\usepackage{url}            
\usepackage{booktabs}       
\usepackage{amsfonts}       
\usepackage{nicefrac}       
\usepackage{microtype}      
\usepackage{lipsum}		
\usepackage{graphicx}
\usepackage{natbib}
\usepackage{doi}

\usepackage{fancyhdr}       
\usepackage{graphicx}       
\graphicspath{{figures/}}     

\date{}

\renewcommand{\headeright}{}
\renewcommand{\undertitle}{}

\renewcommand{\shorttitle}{A Hybrid Model}

\hypersetup{
pdftitle={A Hybrid Model for Forecasting Short-Term Electricity Demand},
pdfsubject={cs.AI},
pdfauthor={Maria Eleni Athanasopoulou, 
Justina Deveikyte, 
Alan Mosca, 
Ilaria Peri and 
Alessandro Provetti
},
pdfkeywords={Hybrid models, Neural Networks, Regression, Feature Engineering},
}

\long\def\COMMENT#1\ENDCOMMENT{\message{(Commented text...)}\par}
\begin{document}
\title{A Hybrid Model for Forecasting Short-Term Electricity Demand}
%

\author{
\href{https://orcid.org/0000-0001-8031-6360}{\includegraphics[scale=0.06]{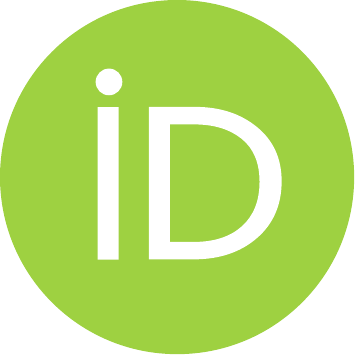}\hspace{1mm}Maria Eleni Athanasopoulou}, 
Justina Deveikyte, 
\href{https://orcid.org/0000-0001-9542-4110}{\includegraphics[scale=0.06]{figures/orcid.pdf}\hspace{1mm}Alan Mosca}, 
\href{https://orcid.org/0000-0001-8918-2389}{\includegraphics[scale=0.06]{figures/orcid.pdf}\hspace{1mm}Ilaria Peri} and 
\href{https://orcid.org/0000-0001-9542-4110}{\includegraphics[scale=0.06]{figures/orcid.pdf}\hspace{1mm}Alessandro Provetti}\\
Department of Computer Science and Information Systems\\
Department of Economics, Mathematics and Statistics\\
Birkbeck - University of London\\
London WC1E 7HX, UK
}

\maketitle             
\begin{abstract}
Currently the UK Electric market is guided by load (demand) forecasts published every thirty minutes by the regulator.
A key factor in predicting demand is weather conditions, with forecasts published every hour.
We present HYENA: a hybrid predictive model that combines feature engineering (selection of the candidate predictor features), mobile-window predictors and finally LSTM encoder-decoders to achieve higher accuracy with respect to mainstream models from the literature.
HYENA decreased MAPE loss by 16\% and RMSE loss by 10\% over the best available benchmark model, thus establishing a new state of the art for the UK electric load (and price) forecasting.
\end{abstract}

\section{Introduction}

Electricity cannot be generated in advance and stored in large quantities to be used at a later time, it needs to be generated in real time. 
Thus, having accurate demand forecasts is crucial to ensuring security of supply.
Furthermore, generators, suppliers and non-physical traders that participate in the electricity market, all need to have an accurate view on future demand.

For this reason, the academic literature examined several methods to improve the accuracy of the demand forecasting, but no model has been proven consistently superior regardless of the dataset and the market. 
Furthermore although the wider topic of load forecasting has been widely researched, there has been significantly less research on electricity systems that have a lot of embedded renewable generation, such as the UK. 

The existence of a significant amount of embedded renewable generation in the system, suppresses load as it covers some of the demand for electricity and as a result less is requested from the grid. 
As the renewable generation is intermittent it introduces more uncertainty to the system, making accurate forecasting more challenging and of higher importance. 
In the UK, there are approximately 30 GWs of embedded generation connected to the distribution system \cite{ng}, making predicting UK national demand extremely challenging. 

Today, the key features that would intuitively be the basis for an electric demand prediction are basically public. 
In particular, past load levels and weather conditions, arguably two key features, are available for extraction from public Web sites. 
Other auxiliary features such as holiday and renewable installed capacity data are also available.

Nonetheless, predicting load on the time-scale of industrial operations, i.e., every thirty minutes, remains a challenge; even well-established methods, including neural networks,incur in huge losses, as shown in our preliminary study on forecasting the current UK electric demand.

In this article we respond to the challenge by defining a new hybrid model, called Hybrid Energy Analyser (HYENA), which we have designed and validated specifically for load forecasting (demand prediction) in the UK electric market.
Our comparative assessment over a long period of historical demand shows that the HYENA model is on average more accurate than any of the established models over which it is based, thus confirming that hybrid models are the best option for this type of forecasting problem.

We approached the challenge by leveraging public data to train a multi-level hybrid system with memory-based machine learning at its core; using traditional forecasting methods to capture seasonal effects and feature engineering to filter the input data; and deploying Machine Learning to the 'internal' problem of learning the deviation in demand from the seasonally-adjusted data.

Our resulting architecture combines Linear regression, to capture seasonality, and a Long-Short Term Memory (LSTM) auto-encoder, to capture the ``residual'' variations in demand.
To validate the proposed model, we used Tao Hong's vanilla model as a benchmark. We also compared the forecasting ability of the hybrid model with that of a wide range of methods such as SARIMA, SARIMA with exogenous variables, Random Forests and simpler LSTM models, using common load forecast accuracy metrics. The proposed model outperformed all other tested methods, achieving a significant error reduction with respect to the benchmark model.

\subsection{Literature on Electric load forecasting}

Academic research has focused on demand forecasting over the years and different forecasting methods have been suggested depending on the forecasting horizon. 
For instance, when forecasting very short-term horizons up to 6 hours, using univariate models such as ARIMA and Holt-Winters exponential smoothing\cite{Taylor2003}, \cite{TAYLOR2008645} is sufficient \cite{TAYLOR20061}, while for longer horizons multivariate models are more appropriate. 
For forecasting up to two weeks ahead, demand is driven by exogenous factors such as weather, calendar and special events. 
Temperature is the main weather-related driver of demand. Its relationship with load is non-linear and it is driven by heating and cooling demand \cite{BENTAIEB2014382}, \cite{HONG2014357}. 
Other factors such as humidity \cite{xie-relative18}, cloud and wind speed \cite{xie-wind17} subtly affect demand and research has shown that using them as predictors improves forecasting accuracy. 
For systems such as California and the UK, where there is a lot of embedded solar generation in the systems suppressing demand, using solar radiation and installed solar capacity as predictors of demand further improves model accuracy \cite{california}. 

One of the most widely used forecasting methods both in academia \cite{hong},\cite{39025} and industry \cite{HONG2014357}, is Multiple Linear Regression (MLR). 
In fact, MLR is able to capture seasonality well with the help of seasonal variables; and can describe the more complex patterns, such as the relationship between load and temperature, using polynomials. 

MLR is highly interpretable and faster to implement than other more complex methods such as Neural Networks. 
Random Forests \cite{lahouar2015} using weather and calendar variables have also been proven to be highly accurate, surpassing the forecasting abilities of other flexible approaches such as Neural Networks.  
Random Forests have the advantage that they can accept large volume of features without majorly impacting their performance when these inputs are not relevant \cite{lahouar2017}. 
Deep learning methods, such as various architectures of LSTM \cite{yu2020} and LSTM hybrid models \cite{tian2018}, \cite{8880605}, have also proven to be promising, as LSTM models are able to learn long-term temporal connections and do not suffer from the vanishing gradient problem. 

Combining LSTM ability to learn long-term dependencies with other models, such as CNN, to capture local trend in data has been shown to outperform simpler LSTM implementations \cite{tian2018}. 
Moreover, \cite{ae} has shown the benefits of using a Stacked Denoising Auto-Encoder, a type of multilayer perceptron (MLP), to learn efficient data coding and use it to accurately forecast demand.

\section{Design and implementation}
The project followed a multi-step process. 
Firstly, we selected the weather stations we would use in our models and gathered the load, weather, holiday and embedded generation installed capacity data. 
In the next stage, we carefully analysed the historic load to uncover the main drivers of the UK electricity demand and engineered features that the analysis showed that could be useful in predicting it. 
We then created our baseline model and other candidate models, applying dimensionality reduction techniques where relevant. 
We fine-tuned the models and, finally, we evaluated them using error metrics commonly used in academia.

\begin{figure}[!htb]
    \centering
    \includegraphics[scale=.5]{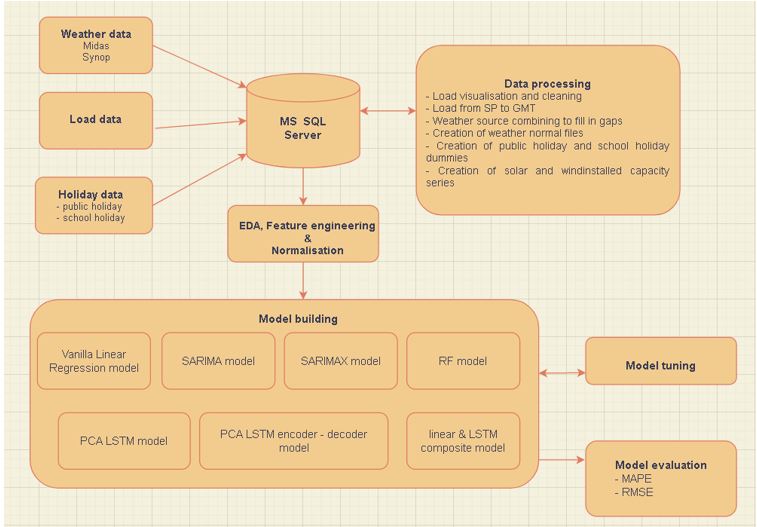}
    \caption{Methodology flowchart}
    \label{fig:proj}
\end{figure}

\subsection{Input data and sources}

\paragraph{Load:} the load data used to build our models was the national demand consumption for the UK and spans between January 2018 and February 2020. 
However, to get a better understanding of the data, its trends and its drivers, we examined the load demand from 2014 until February 2020. 
The data was obtained from the BM Reports API %
which publishes real time data with half hourly granularity.

\COMMENT
in local prevailing time, thus each day has 48 periods, apart from the October and March time change days which have 50 and 46 periods respectively. 
As the electricity market is expressed in settlement periods, we had to shift the data and bring it to GMT timezone, so that it is in line with the timezone the weather was reported in and so as to have a set amount of half hours in a day. 
\ENDCOMMENT

\paragraph{Weather:} the weather data was obtained by the CEDA website using as a primary source the Land Synop data\cite{synop};%
 to complete any missing data we used the Midas open datasets.

\COMMENT
Further gaps in the data were covered by calculating the weather normals for the season using data from January 2014 until May 2020. 
Weather ``normals'' are constructed by calculating the average weather values per month and hour. 
Using weather normals as an input when weather forecasts are not available is a standard practice in the industry, making it a valid way of filling any gaps in the data for this study. 
The variables used for our problem were temperature, wind speed and solar radiation for Heathrow, Coleshill, Bramham, Bishopton and Devon weather stations. 
The reasoning behind these choices will be discussed at a later section of the report.
\ENDCOMMENT

\paragraph{Other inputs}
For the period covered by this study we collected the daily estimation of the embedded solar and wind installed capacity reported by National Grid%
\footnote{Please see the National Grid's \href{https://demandforecast.nationalgrid.com/efs_demand_forecast/faces/DataExplorer}{Data Explorer.}%
}.  
Also, got the exact dates of holidays from the UK government website%
.

\subsection{Data exploration}

\begin{figure}[!htb]
    \centering
    \includegraphics[scale=.8]{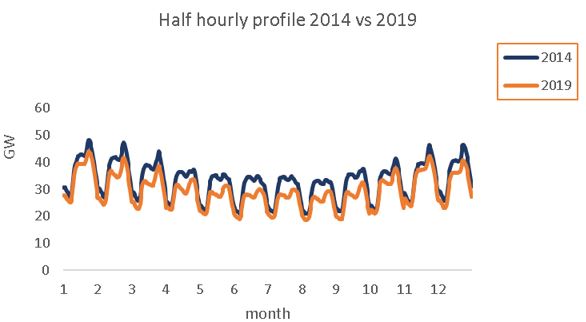}
    \caption{For each month, we plot the average half-hourly load for 2019 (in orange) and 2014 (in blue), the year when renewable installed capacity started its growth.  
    Demand now has a more pronounced dip between 12:00 and 16:00, as a bigger share of demand is covered by solar radiation and less energy is requested from the grid.
    }
    \label{fig:load_profile}
\end{figure}

Looking at the historic electricity demand, it is easy to notice a slow year-on-year decline, shown on Figure \ref{fig:load_profile}, which displays, month by month, the average half-hourly demand. 
The drift is considered a consequence of technological advances, economic factors and the increasing installed capacity of renewable generation. 
To account for this long-term tendency, we used a trend variable in our models.

Examining the historic load at a finer level of detail one can see that the solar and wind installed capacity increase over last few years has in fact changed the shape of demand. 
Between the hours that solar panels can produce substantial amounts of energy (11:00-17:00) we see a dip in demand which is especially pronounced between April and October. 
We also see a dip in the first morning hours due to the embedded wind generating enough power to cover a significant part of the load needed in these low demand hours. 

Given the big impact that embedded renewable generation has had on demand, it is evident that modelling demand requires including some proxy for the embedded generation factor. 
To account for this aspect in the models, at each half-hourly period, we used solar radiation and wind as proxies, while to account for the year on year difference in embedded generation, we used installed capacity as an input.

\noindent
To account for the long-, medium- and short-term trends seen in the data, we had to include features representing both daily, weekly and yearly seasonality.

\COMMENT
Time of the day is an important factor that can help us model demand. 
Examining the load, we see that overnight, demand is low and gradually picks up between 7 and 9 am. 
Around 12:00 we see the first peak of the day, it then dips as a big part of the demand is covered by embedded solar generation and around 6pm we see the evening peak which then gradually drops. 
The data also shows weekly seasonality, thus variables signalling day of week can be useful predictors of demand.

For example, on weekends and holidays the load demand is significantly lower than weekdays and demand on Fridays is slightly lower than the rest of the weekdays. 
Examining the data it is apparent that there is yearly seasonality as well. For example, demand during the warmer months is lower than in the winter months. 
The load profile during the winter months is significantly different than the summer months, with a flatter midday peak and a more pronounced evening peak. 
Thus, variables signalling the month or the season can be useful predictors of demand.

\begin{figure}[!htb]
    \centering
    \includegraphics[scale=1]{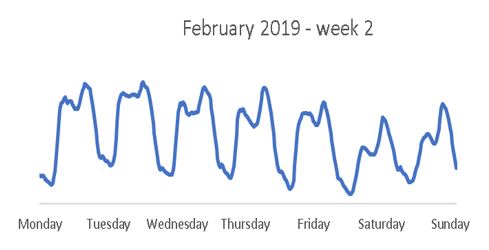}
    \caption{Looking at the half hourly demand of week 2 February 2019 as an example, we see the weekly and hourly seasonality of the load.}
    \label{fig:week}
\end{figure}
\ENDCOMMENT

A further aspect of the load series is its high variance.
In order to ``stabilise'' it over time, we have applied a log transformation on the load, as suggested by \cite{BENTAIEB2014382} before running the benchmark models described below.  
However, the input load to the HYENA model was not transformed.
Vice-versa, as LSTM is sensitive to data scale, the input for the LSTM(-AE) benchmarks described below was normalised to the [0,1] range.
%

\subsection{Weather data}
Weather is one of the main drivers of demand, especially when forecasting more than 6 hours ahead. 
As we set out to model national demand, we included a small set of weather stations that would nonetheless be considered representative of the country's weather. 
One important point is that the stations need to be geographically dispersed so they are not too collinear between them, as collinearities could create unstable load forecasts (please see, e.g., \cite{dormann-collinearity13} for a discussion). 
Keeping this condition in mind, we used population density as proxy for both economic activity in the region and residential energy usage, when we selected the weather stations used in the model. 
Given the significant effect of renewable generation on demand, we also selected weather stations on areas that have high solar and wind installed capacity to act as proxy for the renewable energy being injected to the system.

\COMMENT
\begin{figure}[!htb]
    \centering
    \includegraphics[scale=0.6]{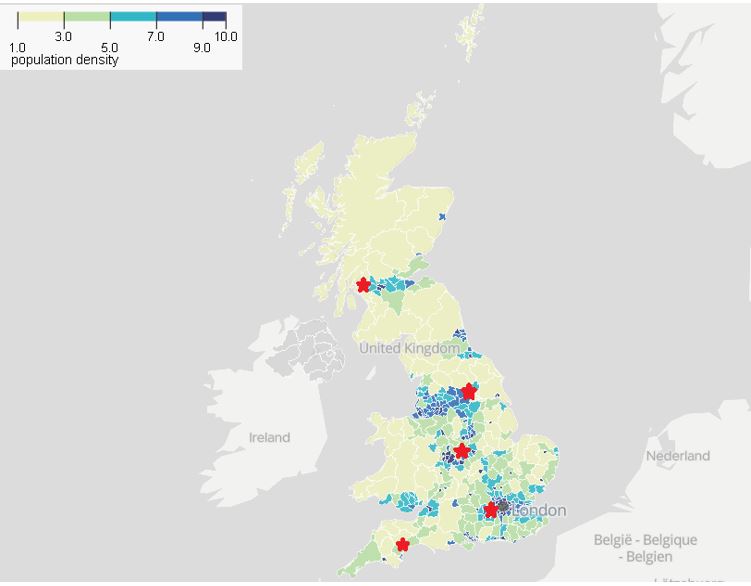}
    \caption{Weather stations selected marked on a population density map. (Source: ONS website)}
    \label{fig:wx}
\end{figure}

\begin{figure}[!htb]
    \centering
    \includegraphics[scale=0.6]{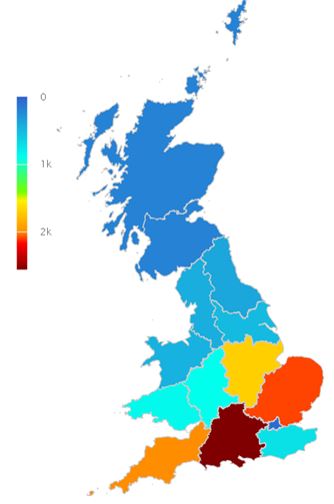}
    \caption{Solar installed capacity map}
    \label{fig:solar}
\end{figure}

\ENDCOMMENT

\subsection{Feature engineering and variable selection}
From the exploratory analysis of the load, we have already discovered a few variables that would be important predictors of demand. 
Demand decreases slightly year on year; thus we included a trend in our models. 
To account to the weekly seasonality, we included a variable for the days of the week. Monthly seasonality is captured with monthly dummies.

\begin{figure}[!htb]
    \centering
    \includegraphics[scale=1]{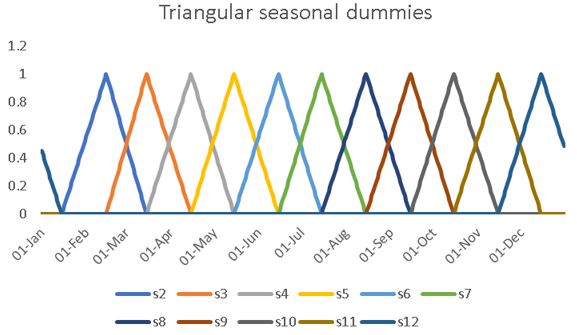}
    \caption{Triangular seasonal dummies}
    \label{fig:triang}
\end{figure}

\begin{figure}[!htb]
    \centering
    \includegraphics[scale=.8]{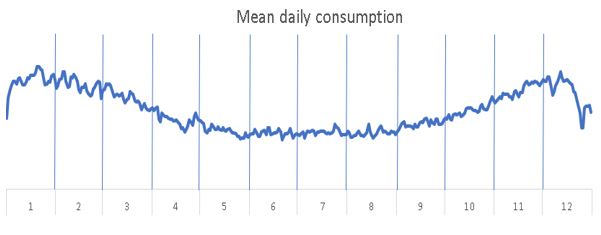}
    \caption{The transition from one month the next is gradual, thus a more sophisticated approach than simple orthogonal dummies is required.}
    \label{fig:ld}
\end{figure}

\COMMENT
As shown in Figure \ref{fig:ld}, the level change from month to month is quite gradual so to better capture this month to month transition, we use monthly dummies.

that would have the same value for the whole month would not be able to capture this effectively. 
For this reason, we used fuzzy logic triangular functions \cite{Smith2007ImprovingAT} ranging from 0 to 1 as shown in Figure \ref{fig:triang} .

\COMMENT
\ENDCOMMENT

To account for the difference in levels between working days and holidays, we included a dummy variable that has the value 1 when the day is a public holiday. 
To capture the daily shape of demand, we included a variable for time. 
Next, we consider the school calendar: as during school holidays we see a later ramp up in demand in the mornings, we added a dummy variable denoting school holidays to our models.

\begin{figure}[!htb]
    \centering
    \includegraphics[scale=.8]{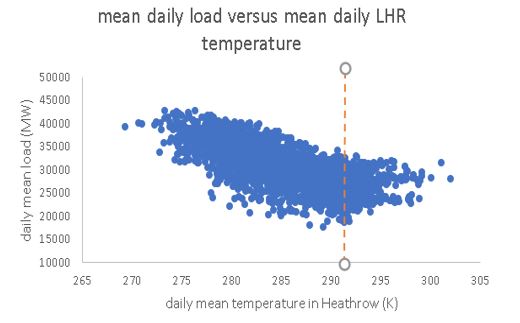}
    \caption{Daily load plotted against mean daily temperature in Heathrow; the non-linear relationship between load and temperature can be observed.}
    \label{fig:temp_curve}
\end{figure}

One of the biggest drivers of demand, beyond seasonality, is temperature. 
Load and temperature have a non-linear relationship. 
As shown in Figure \ref{fig:temp_curve}, when temperature is low, demand increases, while when it is high, demand decreases until it reaches around 291 degrees Kelvin, when the cooling demand starts. 
This can be captured by taking the second-degree polynomial function of temperature.

As we saw in the load exploratory analysis, renewable embedded generation has a big effect on the shape of demand. 
To capture the long-term changes in demand caused by the high renewable penetration, we included variables capturing the installed embedded solar and wind capacity; while to capture the half hourly effect of renewables on load, we included variables showing the solar radiation and wind speed per half hour for the selected 5 weather stations. 
Weather does not have an instantaneous effect on load only. 

\COMMENT
Weather especially temperature and solar, in the previous hours, or even days, has a lingering effect on load. 
To capture these effects, we included lagged weather variables.
\ENDCOMMENT

After we identified the variables that would be good predictors of demand, we used feature importance from Extremely Randomized Regression Trees to validate the variable selection we made based on the exploratory analysis. 
The Extremely Randomized Regression Trees is an ensemble method that fits a number of randomised trees on various sub-samples of the dataset without bootstrapping and uses averaging to improve the predictive accuracy and control over-fitting. 
As expected, the variables selected explained enough of the variance in the load to justify their inclusion in the model; with the most important variables being the time related variables. 
As the UK is not large enough to have multiple climates, we expected some collinearities between weather stations despite purposely selecting them so that they are geographically dispersed.
This was verified by looking at the correlations between stations. 
However, as we are aiming to predict national demand, we chose to include all selected weather stations as we needed to capture weather differences between regions. 
We did however, apply Principal Component Analysis(PCA) in an effort to reduce the dimensions of the dataset and eliminate collinearities in the data. 
PCA identifies the hyperplane that lies closest to the data, projects the data on it, and selects the axis that preserves the maximum amount of variance, so as to lose as little information as possible
%
While it removes any excessive multi-collinearity and reduces data dimensionality, PCA is also known to remove some explainability from data. Since our focus is on prediction, we chose to apply it nonetheless.


\section{The HYENA Model}

HYENA attempts to capture seasonality and trend with a simple linear model, creating a forecast of what the load would be with average weather; next it models the difference between such forecast and the actual observed load. 
This is a novel solution where, instead of trying to predict the signal per se, we first removed the known factors of seasonality and focused on predicting the residual. 

We have implemented HYENA by training the linear model with data between 2014 and 2016 using calendar, installed capacity and holiday data.
For the second step we used an LSTM auto-encoder to forecast the difference between the linear model forecast and the actuals for the period between 2017 and 2019:

\begin{equation}\label{eq:fmp14}
   \varepsilon_t={\hat{y}}_t-y_t
\end{equation}

\noindent
with

\begin{align}
\hat{y}_t = & \beta_0 + \beta_1M_{2t} + {\beta_2M}_{3t} + \beta_3M_{4t} + \beta_4M_{5t} && \nonumber\\ 
   & + {\beta_5SM}_{6t} + {\beta_6M}_{7t} + {\beta_7M}_{8t} + {\beta_8M}_{9t} + {\beta_9M}_{10t} && \nonumber\\ 
   & + {\beta_{10}M}_{11t} + {\beta_{11}M}_{12t} + \beta_{13}t_t + \beta_{14}{wd}_t && \nonumber \\ 
   & + h_{public} + {Trend}_t + C_{solar}+ C_{wind} && \nonumber
\end{align}

\noindent
where $M_{2t}\dots {\ M}_{12t}$ are monthly dummies, $t_t$ is time, ${wd}_t$ is weekday, $h_{public}$ is a public holiday dummy, $C_{solar}$ is the solar installed capacity and $C_{wind}$ is the wind installed capacity.

\begin{figure}[!htb]
    \centering
    \includegraphics[scale=0.8]{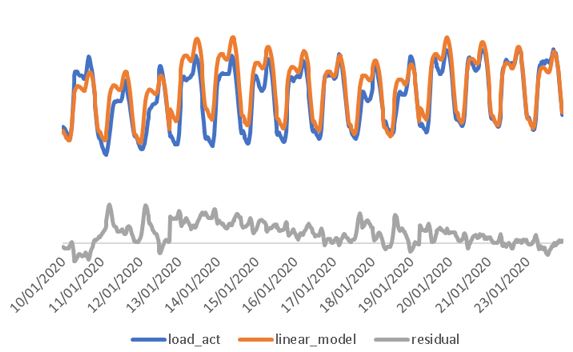}
    \caption{Linear forecast (in orange) to create predictions with average weather and residual demand (in grey) to predict the remaining effects not captured by seasonality and event variables.}
    \label{fig:comp}
\end{figure}

Once the forecast is produced by the linear model, we take its difference from the actual load to create the residual series (Figure \ref{fig:comp}), which represents the deviation from normal demand and is dependent mostly on weather.
To validate this and to select which of the available variables to use in the LSTM auto-encoder model, we used feature importance from Extremely Randomised Trees. 
Using current and lagged variables for temperature and solar radiation, as well as time to capture some remaining seasonality, we built an encoder-decoder LSTM model with an LSTM layer for encoding and an LSTM layer for decoding, followed by a dense layer. 
To optimise the model, we deployed the well-known Adam optimiser with a small batch size and early stopping to avoid over training.

\section{Comparative Model evaluation}

To validate the HYENA approach, and in general to create a comprehensive assessment of predictive models for UK demand forecasting, we have implemented and tuned some of the best-know models from the literature, which are briefly introduced next.
Our baseline model for this prediction task is Tao Hong's ``vanilla'' model, which has been used as a benchmark for the GEFCom2012 energy competition and subsequently in many papers focusing on load forecasting.
Next, we have set up and run instances of 

\begin{itemize}
    \item the Seasonal Autoregressive Integrated Moving Average (SARIMA): an extension of ARIMA that explicitly supports data with a seasonal component, which is the case here;
    
    \item the SARIMAX model, which extends the above by including exogenous variables \cite{Papaioannou_2016} such as weather and calendar variables. 
    SARIMAX thus addresses the potential shortcoming of SARIMA which takes into account past load values;
    
    \item Random Forest, the well-known ensemble method was deployed here by building a random-forest regressor with 200 trees, with maximum depth of 10 and at least 45 samples required to split to an internal node; 
    
    \item LSTM, the well-know deep learning method which can handle non-stationary data (as it is the case with load series) but requires a careful manual tuning.
    Next, Principal component Analysis (PCA) was deployed to reduce the data dimensions down to 9.
    Finally, we implemented a many-to-many LSTM network with 48 timesteps, 9 features and 2 LSTM layers, where each cell per timestep emits a signal to the same timestep of the next layer, followed by a dense layer, and 
    
    \item LSTM autoencoder, which improves LSTM by an  encoding layer which suppresses the inputs and keeps only the most important signals for prediction.
    As we did for LSTM, the model was optimised by minimising the mean square error with the Adam optimiser; also Early stopping was employed to avoid over-training the network.
\end{itemize}

Each model has been tuned, when possible, to best fist historical data and thus make the comparison fair.
For lack of space, we have put the description of this phase in the supplementary material to this article.%

The candidate models were evaluated based on the day-ahead forecasts (i.e., predict the exact level of demand tomorrow). 
Our testing period ran from the 1st of January until the 14th of February 2020. 
Forecasts were evaluated against the retrospective actual values using the Mean Absolute Percentage Error (MAPE) and Root Mean Square Error (RMSE), which are standard for the electricity markets. 
While model training time could be a consideration in this setting, we did not deem it a useful evaluation criterion as none of the models had a prohibitively long training time.

\COMMENT
\begin{table}[!htb]
    \centering
    \begin{tabular}{c|r|r|r|r|r|r|r|}
         & Baseline & SARIMA & SARIMAX & R. Forest & LSTM & LSTM-AE & HYENA \\
        \hline\\
        MAPE & 6.59 & 10.26 & 9.52 & 5.32 & 4.82 & 4.83 & 4.07 \\
        \hline\\
        RMSE & 2468 & 3787 & 3656 & 1965 & 1723 & 1684 & 1517 \\
        \hline
    \end{tabular}
    \caption{Average MAPE and RMSE during the testing period (Jan. 1-Feb. 14 2020).}
    \label{tab:stats}
\end{table}
\ENDCOMMENT

\begin{figure}[!htb]
    \centering
    \includegraphics[scale=1]{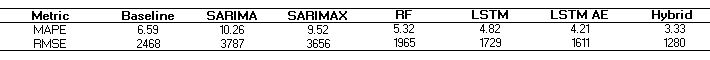}
    \caption{Average MAPE and RMSE during the testing period (Jan. 1-Feb. 14 2020).}
    \label{fig:stats}
\end{figure}

The baseline model (Tao Hong's vanilla model), for all its simplicity, is not easily outperformed, as all models struggled in predicting the ramp-ups in demand. 
However, the more flexible models managed a flatter error shape than the baseline for the rest of the day. 

\begin{figure}[!htb]
    \centering
    \includegraphics[scale=0.9]{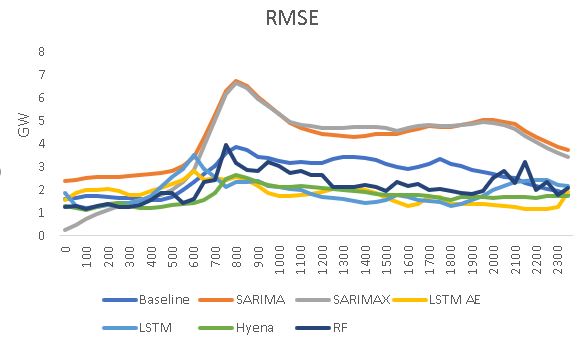}
    \caption{Average RMSE per hour during the testing period (1st Jan to 14th Feb 2020).}
    \label{fig:hh_stats2}
\end{figure}

\noindent
SARIMA and SARIMAX hardly outperformed Hong's baseline model; and contrary to what most of the literature suggests, nor they outperformed LSTMs or Random Forest. 
A possible interpretation, is that our data presentation could have failed to capture the yearly seasonality of the load, as we had to restrict our training sample to rolling 30 days due to restricted computational resources. 

SARIMAX, which takes into account temperature and calendar variables, outperformed SARIMA, but as it was lacking any proxies for solar radiation, it mis-assigned the solar effect leading to higher errors than SARIMA at peak hours for solar radiation (11 am to 3 pm). 
The Random Forest model outperformed the baseline but showed a slightly worse predictive ability than the LSTM models. 

In comparison to the Random Forest, the LSTM model was more difficult to implement, as it required more pre-processing steps and a dimensionality reduction stage while the gains in predictive power over the former where minimal. 
Also the LSTM encoder-decoder implementation managed to outperform all previous models. 
As the LSTM-AE compresses its data, it managed to identify the most important signals in the data, which lead to significantly better performance and less time dependent error.

Finally, our HYENA models outperformed all benchmarks as it captured i) seasonality in data with a well-specified linear sub-model and ii) the residual demand effects with an LSTM encoder-decoder sub-model. 

\section{Conclusions}
Our study of the UK electric market and the multiple factors that affect demand lead us to the design and implementation of HYENA, a hybrid model that shows excellent predictive power: on a long-term dataset containing half-hourly demand data for the first 45 days of 2020 it convincingly outperformed the accuracy or six benchmarks, including statistical (SARIMA, SARIMAX), Machine Learning (Random Forests) and Deep Learning methods (LSTM, LSTM autoencoder): HYENA decreased MAPE by 16\% and RMSE by 10\% over the best available model, thus establishing a new state of the art for the UK electric load forecasting.

HYENA combines existing methods to implement a novel approach to modelling demand, not necessarily electric demand: regression methods to capture seasonality and LSTM autoencoders to capture the ``residual'' variations in demand.
However, several preliminary activities are required to obtain such an effective model. 

A thorough data acquisition and analysis of the load and its drivers was conducted, which highlighted the challenges that are involved when real data is used for model design, such as automatic data collection from multiple inconsistent sources, gaps and temporal inconsistencies in the data, the difficulties of synthesising data from multiple sources to create as a complete a dataset as possible; and unavailability of all the required data. 
Next, a range of forecasting methods was investigated and several intermediate steps, e.g., normalisation and PCA, were executed and evaluated. 
An additional challenge was the limited computational power at our disposal, which was particularly felt for the SARIMAX implementation: more resources would mean a longer training set and a model extension which could account for yearly seasonality. 

Finally, we plan further research on HYENA with a focus on predicting regional demand, which could then be aggregated to create a national demand forecast. 
We expect that this approach would lead to a sharp increase of accuracy, as regional load would be easier to forecast using the relevant weather station, rather than using a few stations to forecast the demand for the whole country. 

This approach would also solve the problem of collinearity
between stations being and would eliminate the need for dimensionality reduction as all of the data would be essential.
Finally, we would be interested in including rain, humidity and cloud data to the proposed model as these would help explain a part of the error which we believe could be significant.

\bibliographystyle{unsrtnat}
\bibliography{hybrid-model}

\end{document}